# Weakly-Supervised Learning-Based Feature Localization for Confocal Laser Endomicroscopy Glioma Images

Mohammadhassan Izadyyazdanabadi[1, 2], Evgenii Belykh[2, 3], Claudio Cavallo[2], Xiaochun Zhao[2], Sirin Gandhi[2], Leandro Borba Moreira[2], Jennifer Eschbacher[2], Peter Nakaji[2], Mark C. Preul[2], and Yezhou Yang[1]

[1] School of Computing, Informatics, and Decision System Engineering, Arizona State University, Tempe AZ 85281, USA
`yz.yang@asu.edu`
[2] Department of Neurosurgery, Barrow Neurological Institute, Phoenix, AZ 85013, USA
[3] Department of Neurosurgery, Irkutsk State Medical University, Irkutsk, 664003 Russia

**Abstract.** Confocal Laser Endomicroscopy (CLE) is novel handheld fluorescence imaging technology that has shown promise for rapid intraoperative diagnosis of brain tumor tissue. Currently CLE is capable of image display only and lacks an automatic system to aid the surgeon in diagnostically analyzing the images. The goal of this project was to develop a computer-aided diagnostic approach for CLE imaging of human glioma with feature localization function. Despite the tremendous progress in object detection and image segmentation methods in recent years, most of such methods require large annotated datasets for training. However, manual annotation of thousands of histopathology images by physicians is costly and time consuming. To overcome this problem, we constructed a Weakly-Supervised Learning (WSL)-based model for feature localization that trains on image-level annotations, and then localizes incidences of a class-of-interest in the test image. We developed a novel convolutional neural network for diagnostic features localization from CLE images by employing a novel multiscale activation map that is laterally inhibited and collaterally integrated. To validate our method, we compared the model output to the manual annotation performed by four neurosurgeons on test images. The model achieved 88% mean accuracy and 86% mean intersection over union on intermediate features and 87% mean accuracy and 88% mean intersection over union on restrictive fine features, while outperforming other state of the art methods tested. This system can improve accuracy and efficiency in characterization of CLE images of glioma tissue during surgery, and may augment intraoperative decision-making regarding the tumor margin and improve brain tumor resection.

## 1    Introduction

Rapid intraoperative interpretation of suspected brain tumor tissue is of paramount importance for planning the treatment and guiding the neurosurgeon towards the op-timal extent of tumor resection. Handheld, portable Confocal Laser Endomicroscopy (CLE) is being explored as a fluorescence imaging technique for its ability to image histopathological features of tissue at cellular resolution in real time during brain tumor surgery [1–4]. CLE systems can acquire up to 20 images per second, with areas in the tumor resection bed interrogated as an "optical biopsy". Hundreds of images may be acquired showing thousands of cells, but the images may be affected with artifacts such as red blood cells (for CLE systems operating in the blue laser range) and motion distortion, making them complicated to analyze. Although images may be interpreted as largely artefactual, detailed inspection often reveals image areas that may be diagnostic. CLE images present a new fluorescent image environment for the pathologist. Augmenting CLE technology with a computer aided system that can rapidly highlight image regions that may reveal malignant or spreading tumor would have great impact on intraoperative diagnosis. This is relevant for tumors such as gliomas where discrimination of margin regions is key to achieve maximal safe resection, which has been correlated with increased patient survival duration [5, 6].

Recent studies have shown that off-the-shelf Convolutional Neural Networks (CNNs) can be used effectively for classifying CLE images based on their diagnostic value [7, 8] and tumor type [9]. However, feature localization models have not been previously applied to CLE images. Feature localization models based on fully supervised learning require large number of images for object-level annotation of the features, which is expensive and time consuming. To overcome this limitation, we used a weakly-supervised localization (WSL) approach. A WSL approach allowed the model to learn and localize the class-specific features from image-level labels.

A few groups have recently applied WSL approaches to medical images, including placenta scans [10], whole-slide images of colorectal cancer [11], diabetic retinopathy [12], microscopic cellular images [13], and lung computed tomography scans [14]. Here, we present a novel model for detection of histological features of glioma on CLE images trained on a dataset of CLE images acquired during brain surgery for this invasive tumor. The architecture included end-to-end Multi-Layer Class Activation Map (MLCAM) with Lateral Inhibition (LI) and Collateral Integration (CI) of the glioma feature localizer neurons. The model was able to segment the CLE images semantically by disentangling class-specific discriminative features that can complement interpretation by the physicians. Performance of the model was assessed by comparing its output to CLE image segmentations performed by neurosurgeons and other deep learning models. Additionally, we validated the significance of the MLCAM, LI and CI architecture components on the overall performance of the model. The model localized known diagnostic CLE features and revealed new CLE features that correlated with the final classification and were not previously recognized by the reviewers.

Unlike previous models that require patch labeling [11] or an extra step for creating the activation maps during testing [15], our model is solely trained based on the whole image-level labels. Furthermore, we did not limit the network to localize features that are already known phenotypes to the physicians [13, 14]. CLE images are relatively novel to the pathology tissue diagnosis workflow. Although the tissue architecture suggestive for a certain tumor type can be identified on CLE images [1–4], detailed characteristic brain tumor patterns for CLE images are not yet well described. Therefore, we used a more general concept (glioma diagnostic vs. nondiagnostic) that includes a range of known histological diagnostic elements (i.e., large nucleus, mitotic figures, hypercellularity, etc.) and allows for discovery of previously unrecognized features that may correlate with final image classification. Further investigation of detected features may deepen the understanding of glioma histopathological phenotypes in CLE images, consequently improving their theranostic implications.

## 2 Methods

We constructed a WSL-based model to generate glioma Diagnostic Feature Maps (DFM) from CLE images, which includes three main components (see Figure 1): 1) Customized CNN architecture with new design of CAM at different CNN layers. 2) Lateral inhibition (LI) mechanism that suppresses the activation of DFM at locations where its competitor, nondiagnostic feature map (NFM), also exhibit high activation. 3) Collateral integration (CI) mechanism that amplifies activation of DFM at locations where its allies at other layers also have high activations.

For an input image $I_m$ supplied to the CNN, the class scores ($S_D$ for diagnostic and $S_N$ for nondiagnostic) are defined from three layers via global pooling of discriminative regions estimated in each activation map (DFM, NFM). The class scores achieved from each layer, are then passed to independent softmax layers. The three predictions (probability of $I_m$ being diagnostic (D) and nondiagnostic (ND)) achieved from the softmax layers are streamed into three multinomial logistic loss layers and inject the weight update into the CNN during backpropagation. The total loss is calculated by summing the three loss values.

### 2.1 New Design of Class Activation Map (CAM)

To produce the CAM from each layer, a new convolutional layer is stacked to sum its weighted feature planes. Formally, the DFM and NFM at location $(x, y)$ achieved from layer $z^j$, are defined as:

$$DFM(x, y, z^j) = \sum_l w_{k^1}^{z^j} f_l(x, y, z^j), \tag{1}$$

$$NFM(x, y, z^j) = \sum_l w_{k^0}^{z^j} f_l(x, y, z^j), \tag{2}$$

where $f_l(x, y, z^j)$ is the activation of $l^{th}$ feature plane of layer $z^j$ at location (x, y) and $w_{k^1}^{z^j}$ and $w_{k^0}^{z^j}$ are the weights to produce the DFM and NFM, respectively. By applying GAP and then softmax function on DFM and NFM, the classification scores for different classes are calculated at each layer. Therefore, the softmax input for diagnostic ($S_D$) and nondiagnostic ($S_N$) class at layer $z^j$ can be formulated as:

$$S_D = \frac{1}{W^{z^j} \times H^{z^j}} \sum_{x,y} \text{DFM}(x, y, z^j) = \frac{1}{W^{z^j} \times H^{z^j}} \sum_{x,y} \sum_l w_{k^1}^{z^j} f_l(x, y, z^j), \tag{3}$$

$$S_N = \frac{1}{W^{z^j} \times H^{z^j}} \sum_{x,y} \text{NFM}(x, y, z^j) = \frac{1}{W^{z^j} \times H^{z^j}} \sum_{x,y} \sum_l w_{k^0}^{z^j} f_l(x, y, z^j), \tag{4}$$

where $W^{z^j}$ and $H^{z^j}$ are the width and height of DFM and NFM at layer $z^j$. With the novel design of MLCAM, DFM, and NFM are produced in every forward pass and are updated through backpropagation. Furthermore, producing DFM from deeper layers empowers the overall predictive power of the model (i.e. labeling the detected region as diagnostic or nondiagnostic), while DFM from shallower layers allows larger spatial resolution and more precise detection of fine regions.

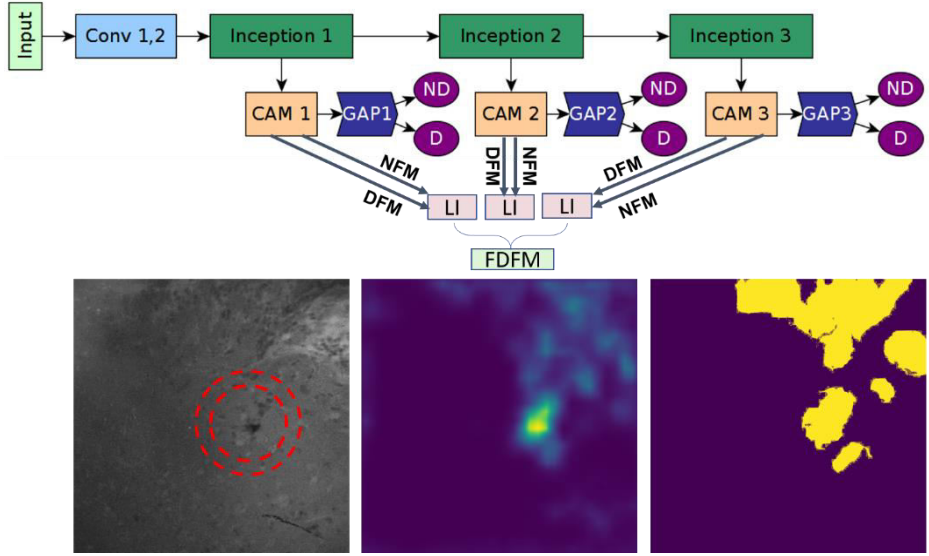

**Fig. 1.** Network architecture with Lateral Inhibition (LI) and Collateral Integration (CI) components for weakly supervised localization of glioma diagnostic features. Bottom image shows a CLE image along with the final diagnostic feature map generated by the model.

### 2.2 Lateral Inhibition and Collateral Integration of Localizer Neurons

During the computation of DFM and NFM, some locations might be activated in both feature maps, which indicates the model's confusion about the diagnostic value of those regions. The activation of DFM is downregulated in these regions, using NFM activations. This mechanism is known as neuronal lateral inhibition in neurobiology [16]). Furthermore, we upregulate the activation of regions which had higher recurrence of activation by integrating DFMs achieved from different layers. To combine these two neural interactions, we compose the following equation to produce the Final DFM (FDFM):

$$FDFM(x,y) = \sum_{z^i, z^j (i \neq j)} [DFM'(x,y,z^i) - DFM'(x,y,z^i).NFM'(x,y,z^i)].[DFM'(x,y,z^j) - DFM'(x,y,z^j).NFM'(x,y,z^j)], \quad (5)$$

where $DFM'(x,y,z^i)$ and $NFM'(x,y,z^i)$ are the value of normalized diagnostic and nodiagnostic feature maps achieved from layer $z^i$, after up-sampling to the original input image size. As shown in Eq. (5), the downregulation for layer $z^i$ is implemented by subtracting the $DFM(x, y, z^i).NFM(x, y, z^i)$ term, which represents the confusing regions at this layer, from $DFM(x, y, z^i)$. Lastly, $FDFM(x, y)$ is also normalized. Figure 1 presents the developed network's architecture. The three inception modules have the same architecture, each combines filters of size 1 × 1, 3 × 3, 5 × 5 in parallel, and concatenates the outputs from each filter into a single tensor [17].

## 3    Experimental Setup and Results

To train our model on image-level annotations, first, a "classification dataset" was created. The CLE images were acquired with an Optiscan 5.1 CLE as described previously [1]. The classification dataset included 6,287 CLE images (3,126 diagnostic and 3,161 nondiagnostic) from 20 patients with glioma brain tumors. If the CLE image depicted any distinguishable diagnostic features, it was labeled as diagnostic and otherwise as nondiagnostic. Table 1 shows the composition of the classification dataset and the number of images used in each stage.

**Table 1.** Number of Diagnostic (D) and Nondiagnostic (ND) images used for training, validation (Val), and test stage is presented.

|       | D    | ND   | All  |
|-------|------|------|------|
| Train | 1714 | 1729 | 3443 |
| Val   | 487  | 511  | 998  |
| Test  | 925  | 921  | 1846 |
| Total | 3126 | 3161 | 6287 |

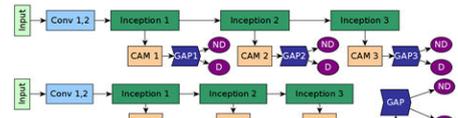

**Fig. 2.** Network architectures used for the ablation study. Top network shows the developed architecture without the LI and CI components. Bottom network shows the MLGAP architecture [14] which combines the three CAMs and then uses a GAP layer for classification.

The classification dataset was divided on a patient level for model development and test (12 cases for training, 4 cases for validation and 4 cases isolated for testing). Stochastic Gradient Descent (SGD) with an initial learning rate of 0.001 and momentum of 0.9 was used to optimize the model's parameters. Learning rate decay policy was set to step function with a gamma of 0.9 and step size of 500 iterations. Image cropping and rotation were not used for augmentation because these might harm the validity of images. Since the diagnostic features could be very small, not every crop of a diagnostic image would be diagnostic. Also, there is no guarantee that the acquired CLE images are rotation invariant (e.g. the surgeons' preference for holding the CLE probe). Training batch size was set to 15 images and it took 22,000 iterations to achieve the model with the minimum loss on classification of validation images. All the experiments were performed in Caffe [18] deep learning framework, using a GeForce GTX 980 Ti GPU (6 GB memory).

The classification accuracy of the model was 84% on the test set (sensitivity = 83.8%, specificity = 84.1%). To validate the efficacy of the WSL model, we tested the following three hypotheses. *First*, the model can correctly segment the image regions which have features that are indicative of glioma, confirmed by physicians at different scales (i.e., medium-sized intermediate and small-sized restrictive scales) and without much reliance on previous exposure (i.e., images from training, validation and test stages). *Second*, the new components utilized (MLCAM, LI, and CI) increase the performance of the model in detecting the features (especially restrictive features) compared to the other state of the art WSL methods that lack them and removing any of these would affect the model performance negatively. *Third*, the developed method can detect novel features in CLE images that were not previously recognized by the physicians. The three hypotheses were tested empirically, using image semantic segmentation task with the following evaluation metrics: mean accuracy (mean_acc), mean intersection over union (mean_IU), and frequency-weighted intersection over union (fw_IU).

A segmentation dataset including 310 CLE images was acquired from images annotated by four neurosurgeons. Each observer highlighted the diagnostic glioma features of each CLE images, independently. We used majority voting to process the annotation variations from the neurosurgeons. For rigorous assessment of the first hypothesis, the segmentation dataset included diagnostic regions at different scales. (145 images were annotated for both Intermediate (Set2-I) and Restrictive (Set2-R) features). Also, to study the effect of previous exposure of CLE image to the model, we used images from all three stages: 30 images from training (Set1), 145 images form validation (Set2), and 135 images from test set (Set3 and Set4)). To appraise the second hypothesis, we sequentially altered components of the designed architecture and assessed the resulting performance of the model ("ablation study"). All models were trained and tested on the same data with the same parameters to avoid any bias. Finally, to test the third hypothesis, our dataset included 55 CLE images that were known to be from glioma tumors but were initially classified as nondiagnostic (Set4). The model generated the segmentation mask by creating the FDFM of the input image with one forward pass and then thresholding (threshold value of 0.03 for intermediate and 0.2 for restrictive features).

Table 2 shows experimental results of segmentation performance by ten different models with respect to the annotators. Each model constructs a DFM to create a segmentation map: M1, similar to [14]; M2 – DFM and NFM of CAM 1,2, and 3 are first laterally inhibited and then collaterally integrated; M3 – CAM 1,2, and 3 are collaterally integrated; M4, M5, M6 – by laterally inhibiting the DFM and NFM of CAM 1, 2, and 3, respectively; M7, M8, M9 – by using the DFMs from CAM 1, 2, and 3 without any further processing; M10, similar to [15]. The first hypothesis proved to be true, since our developed model, M2, produced high mean_acc, mean_IU, and fw_IU for all the intermediate features from diagnostic images (Set1, Set2-I, and Set3). Moreover, it could segment the images from Set3 without significant change in mean_acc, while producing better fw_IU and mean_IU values on images that were previously revealed to it (Set1). Results from Set2-I and Set2-R images showed that all models generated much lower mean_IU and fw_IU on restrictive features compared to intermediate features, except for M1 and M2 models, both of which utilize shallower layers for enhancing the DFM's spatial resolution. In all experiments, M2 made the best performance for three measures (except in mean_acc for Set2-R), supporting the second hypothesis about the significance of the utilized components (MLCAM, LI, and CI). Specifically, M4-M6 models outperformed other ablated models (M7-M9), highlighting the significant value of LI. The higher mean_IU value of M6 and M9 compared to M4,5 and M7,8, respectively, indicates that more abstract features were learned by inception 3 than by inception 1,2. In the first round of review, clinicians labeled Set4 images as nondiagnostic, however, after features were highlighted by the developed model, the clinicians re-classified Set4 images as diagnostic. The highest performance in Set4 belonged to M2 (mean_acc = 88% and mean_IU = 89%). High mean_IU value achieved by the model and clinical feedback emphasize significance and novelty of the features.

**Table 2.** Segmentation performance by different models. M2* is the developed model.

|  | Set | M1 | M2* | M3 | M4 | M5 | M6 | M7 | M8 | M9 | M10 |
|---|---|---|---|---|---|---|---|---|---|---|---|
| mean_acc | Set1 | 0.71 | **0.88** | 0.71 | 0.75 | 0.75 | 0.77 | 0.71 | 0.71 | 0.71 | 0.7 |
|  | Set2-I | 0.76 | **0.85** | 0.74 | 0.76 | 0.76 | 0.77 | 0.74 | 0.74 | 0.74 | 0.74 |
|  | Set3 | 0.72 | **0.86** | 0.72 | 0.75 | 0.75 | 0.76 | 0.72 | 0.72 | 0.72 | 0.72 |
|  | Set2-R | 0.78 | 0.87 | 0.79 | **0.88** | **0.88** | 0.85 | 0.78 | 0.79 | 0.81 | 0.78 |
|  | Set4 | 0.74 | **0.88** | 0.74 | 0.76 | 0.76 | 0.78 | 0.74 | 0.74 | 0.74 | 0.72 |
| mean_IU | Set1 | 0.65 | **0.9** | 0.61 | 0.69 | 0.69 | 0.72 | 0.61 | 0.61 | 0.61 | 0.63 |
|  | Set2-I | 0.69 | **0.86** | 0.67 | 0.69 | 0.71 | 0.73 | 0.65 | 0.67 | 0.67 | 0.69 |
|  | Set3 | 0.57 | **0.82** | 0.56 | 0.59 | 0.61 | 0.63 | 0.56 | 0.56 | 0.56 | 0.59 |
|  | Set2-R | 0.77 | **0.88** | 0.29 | 0.57 | 0.63 | 0.59 | 0.27 | 0.29 | 0.31 | 0.63 |
|  | Set4 | 0.48 | **0.89** | 0.48 | 0.52 | 0.55 | 0.57 | 0.48 | 0.48 | 0.48 | 0.5 |
| fw_IU | Set1 | 0.8 | **0.99** | 0.8 | 0.83 | 0.85 | 0.87 | 0.8 | 0.8 | 0.8 | 0.8 |
|  | Set2-I | 0.88 | **0.98** | 0.86 | 0.88 | 0.9 | 0.92 | 0.86 | 0.86 | 0.86 | 0.86 |
|  | Set3 | 0.65 | **0.88** | 0.65 | 0.69 | 0.71 | 0.73 | 0.65 | 0.65 | 0.65 | 0.67 |
|  | Set2-R | 0.9 | **0.97** | 0.18 | 0.5 | 0.61 | 0.58 | 0.14 | 0.16 | 0.2 | 0.67 |
|  | Set4 | 0.38 | **0.79** | 0.35 | 0.42 | 0.44 | 0.46 | 0.35 | 0.35 | 0.35 | 0.4 |

## 4 Conclusions

In this study, a WSL model was developed to localize the diagnostic features of gliomas in CLE images. It utilizes three fundamental components for creating the final glioma DFM: multi-scale DFM, LI for removing confusing regions, and CI to spatially infuse diagnostic areas from DFMs with different spatial resolutions. The model could detect the diagnostic regions with high agreement compared with annotation by neurosurgeon, from both diagnostic and nondiagnostic images (i.e., images that were initially designated as lacking diagnostic features) in intermediate and restrictive features, while outperforming other methods. Such an approach should be tested on larger datasets. Initial testing demonstrated that WSL has the potential to identify not only relevant, but novel or unrecognized diagnostic features in CLE images that were not previously discriminated by human inspection, requiring further investigation. This approach can be augmented with active learning and patch clustering to create an atlas of glioma phenotypes in CLE images. Further detailed studies correlating regular histology and CLE images are necessary for better understanding of glioma histopathological features on CLE images.

Acknowledgement: YY is partially supported by NSF grant #1750802. This work was supported by the Newsome Chair in Neurosurgery Research held by MCP and by funds from the Barrow Neurological Foundation. EB acknowledges SP-2044.2018.4.